\documentclass[review]{elsarticle}

\usepackage{lineno,hyperref}
\modulolinenumbers[5]

\journal{Neural Networks}

\newcommand{\etal}{\mbox{\textit{et al. }}}

\usepackage[cmex10]{amsmath}
\usepackage{epstopdf}
\usepackage{algorithmic}
\usepackage[ruled,vlined]{algorithm2e}
\usepackage{url}
\usepackage[dvipsnames]{xcolor}
\usepackage{multirow}
\usepackage{hhline}
\usepackage[shortlabels]{enumitem}
\usepackage{pbox}
\usepackage{rotating}
\usepackage[export]{adjustbox}
\usepackage[skip=0pt]{caption}
\usepackage{booktabs}
\usepackage{amsmath}
\usepackage{amssymb}
\usepackage{subfigure}

\biboptions{sort&compress}

\begin{document}

\begin{frontmatter}

\title{Visual Question Answering based on Local-Scene-Aware Referring Expression Generation}

%
\author[add2]{Jung-Jun Kim\fnref{fn1}}
\author[add4]{Dong-Gyu Lee\fnref{fn1}}
\fntext[fn1]{Equally contributed}
\author[add3]{Jialin Wu}
\author[add2]{Hong-Gyu Jung}

\author[add1]{Seong-Whan Lee\corref{corauth}}
\cortext[corauth]{Corresponding author}
\ead{sw.lee@korea.ac.kr}

\address[add2]{Department of Brain and Cognitive Engineering, Korea University, \\Anam-dong, Seongbuk-gu, Seoul 02841, Korea}
\address[add4]{Department of Artificial Intelligence, Kyungpook National University,\\ Daehak-ro, Buk-gu, Daegu 41566, Korea}
\address[add3]{Department of Computer Science, University of Texas at Austin, \\ Austin, TX 78712, United States}
\address[add1]{Department of Artificial Intelligence, Korea University,\\ Anam-dong, Seongbuk-gu, Seoul 02841, Korea}

%

\begin{abstract}
Visual question answering requires a deep understanding of both images and natural language. However, most methods mainly focus on visual concept; such as the relationships between various objects.
The limited use of object categories combined with their relationships or simple question embedding is insufficient for representing complex scenes and explaining decisions.
To address this limitation, we propose the use of text expressions generated for images, because such expressions have few structural constraints and can provide richer descriptions of images.
The generated expressions can be incorporated with visual features and question embedding to obtain the question-relevant answer. A joint-embedding multi-head attention network is also proposed to model three different information modalities with co-attention. We quantitatively and qualitatively evaluated the proposed method on the VQA v2 dataset and compared it with state-of-the-art methods in terms of answer prediction. The quality of the generated expressions was also evaluated on the RefCOCO, RefCOCO+, and RefCOCOg datasets. Experimental results demonstrate the effectiveness of the proposed method and reveal that it outperformed all of the competing methods in terms of both quantitative and qualitative results.
\end{abstract}


\end{frontmatter}

\section{Introduction}
\label{sec:intro}
Over the past few years, visual question answering (VQA) has attracted substantial attention from both the computer vision and natural language processing communities \cite{anderson2018bottom, fukui2016multimodal, ben2017mutan, antol2015vqa, goyal2017making, yu2019deep, teney2018tips, lu2018r}.
Compared to the traditional tasks of computer vision or natural language processing, such as object detection \cite{ren2015faster}, image captioning \cite{rennie2017self, lu2017knowing, pedersoli2017areas, liu2017referring, yu2017joint}, tracking \cite{roh2007accurate, park2013face}, face recognition \cite{maeng2012nighttime, kang2014nighttime}, action recognition \cite{roh2010view, kim2020three, lee2019prediction}, or machine translation \cite{cho2014learning, bahdanau2014neural}, the VQA is a challenging task that requires a more fine-grained semantic understanding of both questions and images jointly, as well as common sense knowledge to answer accurately.
The recently collected VQA v2 dataset \cite{antol2015vqa, goyal2017making} provides complementary pairs of questions and answers.
Each pair includes the same question for two semantically similar images, but the corresponding answers are different.
Because the two images are semantically similar, a VQA model must be able to perform fine-grained reasoning by understanding the details of the scene context to answer questions correctly.
In this paper, we propose a framework that can answer visual questions about a given image and generate human-interpretable textual explanations for answers.

Most VQA research focuses on directly exploiting visual features, such as object attributes or the relationships between objects, to understand images \cite{lu2018r, anderson2018bottom, pedersoli2017areas, lu2017knowing, hong2019exploiting}.
Some studies have utilized image captions or descriptive paragraphs to represent the semantic meaning of the image. The semantic understanding with higher-level information of the scene such as contexts and relationships which are difficult to directly extract from low-level features, can be inferred through the descriptive paragraphs. Such textual representations can contain the richer and necessary information to improve model performances \cite{wu2019generating, kim2019improving}.
Appropriate captions are highly helpful for VQA, which is demonstrated by the fact VQA models that employ human-annotated captions without images can outperform many VQA models that use image features \cite{qingli_tell_answer}.

However, representing an image using a single sentence has limitations in terms of expressional power, and it is easy to overlook important attributes or objects when images become complex or when a question asks a model for specific information. Considering a few specific objects with positive pairs yields poor precision and can often generate non-existent objects in complex scenes \cite{liu2017referring, mao2016generation, tanaka2019generating, yu2016modeling, yu2017joint}.

To address this issue, we propose a novel speaker module that can produce referring expressions for each specific object. This module is tailored for answer prediction and is aware of local contextual information. The referring expression for a target object, which is a question-relevant object in an image, is generated by considering contextual spatial neighbors. The speaker module can describe detailed local scenes, allowing a VQA model to learn target-object-relevant answers.

\begin{figure}[!t]
\centering
\includegraphics[width=1\linewidth]{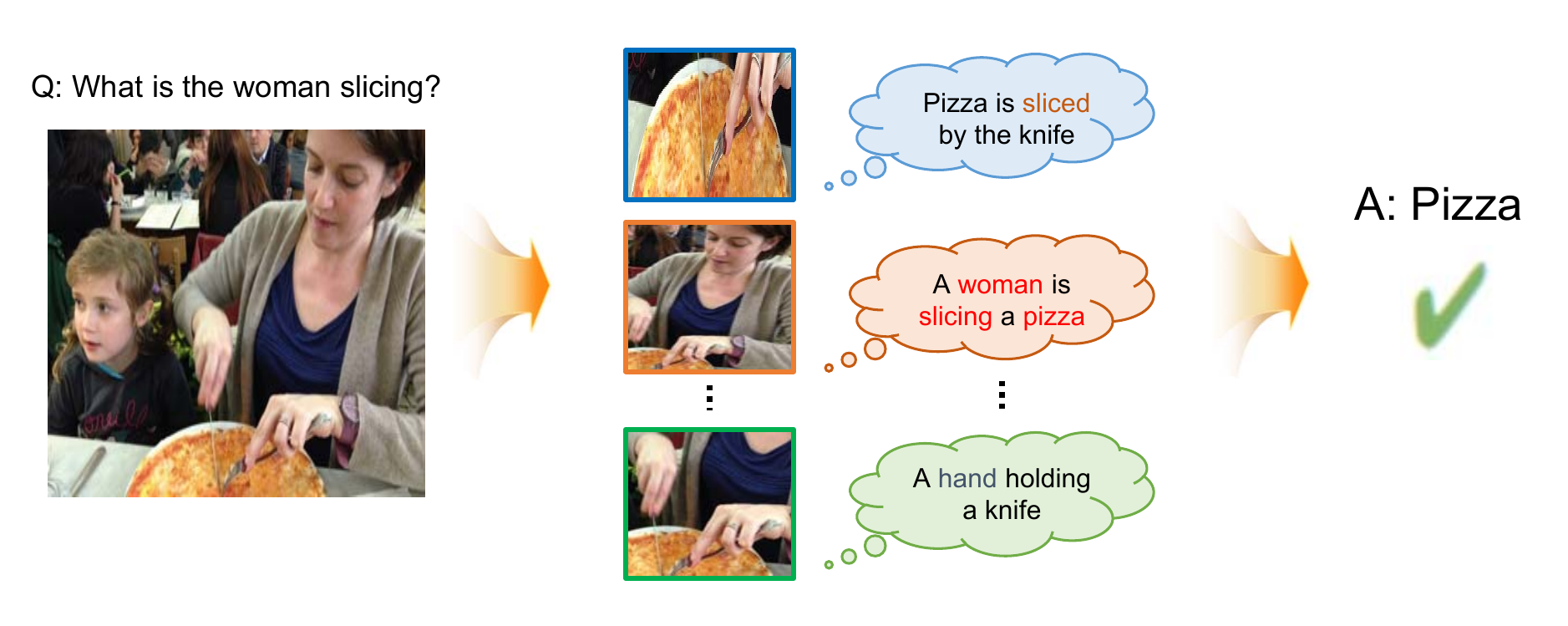}
\caption{Visualization of the concept of our proposed approach to VQA. Multiple object expressions for an image are generated using our local-scene-aware speaker module. Then, the VQA model answers questions based on question-relevant object expressions using one of three input modalities.
}
\label{fig:overview}
\end{figure}

As shown in Fig. \ref{fig:overview}, an image can be represented by multiple text expressions for each detected object from object detection. The relationships between multiple objects are efficiently expressed by text. The proposed method predicts the answer ``pizza'' by simultaneously considering text expressions ``woman slicing pizza'' as well as the attributes ``woman'', ``slicing'', and ``pizza''. The generated expression ``a woman is slicing a pizza'' serves as the foundation for the final answer.
Attention mechanisms have been applied in numerous uni-modal deep neural networks \cite{mnih2014recurrent, bahdanau2014neural, shih2016look}.
In recent studies, the simultaneous learning of co-attention for visual and textual modalities has yielded fine-grained representations of images and questions for VQA tasks \cite{anderson2018bottom, lu2017knowing, fukui2016multimodal}. Inspired by the Transformer model \cite{vaswani2017attention} and Co-attention networks \cite{yu2019deep, nguyen2018improved, kim2018bilinear}, we designed a joint-embedding multi-head attention (JE-MHA) network, consisting of a combination of three multi-head attention (MHA) modules for learning multi-modal embeddings.
The JE-MHA network takes three types of inputs: visual feature, question and text expressions of an image. Because generated expressions are question-relevant representations, owing to the representational power of the speaker module, a specific semantic understanding of a target image associated with the provided questions allows a VQA model to learn the correct answers to questions.

Overall, our primary contributions can be summarized as follows: (1) we propose a novel speaker module that can contextually understand target objects by considering neighboring objects. This allows us to generate local-scene-aware object text expressions. (2) we designed a VQA framework with the proposed JE-MHA network; this framework can model the question-relevant text expressions with images and questions. (3) the answer prediction performance of the proposed VQA model was validated with complementary pairs. The quality of the generated explanations was also evaluated and compared to that of other methods. The results indicate that the proposed model focuses on important factors for answer prediction.

The remainder of this paper is organized as follows. In Section 2, we discuss studies related to our work. In Section 3, we provide the details of the proposed methods for VQA and expression generation. In Section 4, we present an analysis of the results of our experiments. Finally, in Section 5, we conclude this paper.

\section{Related Work}
\label{sec:related}

\subsection{Visual question answering}
Recent studies on deep learning based VQA systems have used bottom-up visual features \cite{anderson2018bottom, wu2019generating,yu2019deep,qingli_tell_answer, kim2019improving}. Such systems detect object attributes using pre-trained detectors, trained on the noisy labeled Visual Genome dataset \cite{krishna2017visual}.
Object-level features help VQA systems attain a semantic understanding of images. One successful approach to VQA is the Transformer model \cite{vaswani2017attention} with Co-attention \cite{yu2019deep, nguyen2018improved, kim2018bilinear}. These methods employ a pre-trained text corpus that was collected from a larger corpus in the natural language processing field. A large number of VQA systems have adopted similar ideas of combining attention over visual features and language features to increase model capacity for question-relevant image understanding \cite{fukui2016multimodal, lu2016hierarchical, xu2016ask}.

Meanwhile, more recently, some researchers have adopted text information as an input modality for VQA, such text is generated from images to supplement the visual relationships between object representations \cite{qingli_tell_answer, wu2019generating, kim2019improving}.
Li \etal \cite{qingli_tell_answer} used a pre-trained captioner to generate general captions based on a fixed annotator.
The generated captions for image were fed into an answer predictor in a VQA system. Therefore, these captions were not necessarily relevant to the target questions.
Wu \etal \cite{wu2019generating} proposed a question-relevant image captioner to generate captions that are more likely to help the VQA system answer questions. A novel caption-embedding module recognized important words in captions, and produced caption embeddings tailored for answer prediction.
Kim \etal \cite{kim2019improving} proposed a visual and textual question answering model to generate caption paragraphs that describe image details to use text as an additional input feature. Their model combines information from text and images via early fusion, late fusion, and later fusion.
However, when a scene becomes complex, image captions can easily disregard important objects. Therefore, it is difficult to represent such scenes accurately. To address this issue, we propose the JE-MHA network, which can consider image, question and expression embeddings, simultaneously. The proposed speaker module generates referring expressions considering local-scene-aware neighboring objects. Common contexts of targets can help identify important objects and their relationships in a complex scene.

\subsection{Referring expression generation}
Referring expression generation is the task of generating an informative sentence for a particular object in a complex scene. Therefore, referred sentences are more unique and specific than general image captions.
Most studies on referring expressions generation have directly compared the difference of visual attributes, positions and sizes between the target object and positive pairs in the scene \cite{liu2017referring, mao2016generation, tanaka2019generating, yu2016modeling, yu2017joint}. The positive pair means that the target object and its pair object belong to the same category, usually the most relevant object within an image, while the negative pair means that the pair object has different categories from the target object \cite{yu2016modeling}.
Such methods generate sentences regarding particular objects while reducing the ambiguity of expressions between multiple objects in a scene. Therefore, large-scale referring expression generation datasets (RefCOCO ,RefCOCO+ , RefCOCOg) have been developed, for this task \cite{kazemzadeh2014referitgame}.
Mao \etal \cite{mao2016generation} introduced a max-margin maximum mutual information (MMI) training method that solves the problem of sentence ambiguity and improves the performance of both generation and comprehension tasks.
Their method improved expressive power for both generation and comprehension tasks.
Yu \etal \cite{yu2016modeling} proposed an explicit encoding method that compares visual feature between objects of the same category in an image. They focused on the most salient objects, rather than general positive pair objects.
Liu \etal \cite{liu2017referring} explored the role of attributes of objects and their paired descriptions. Expressions were generated by incorporating visual and learned attributes into the same semantic space.
Yu \etal \cite{yu2017joint} proposed a unified framework for both referring expression generation and comprehension tasks. A speaker module and listener module were jointly trained by minimizing the distance between paired objects and sentences mapped in an embedding space.
Tanaka \etal \cite{tanaka2019generating} proposed a method that can be interpreted accurately and quickly by humans. When a target is not salient, their model generates expressions by extracting information from the target and environment. A delicate understanding of scene context is essential for referring expression generation. In this paper, the representational power of detailed object relationships was maximized based on the referring expressions of each target object. This method helps a VQA model understand the context of complex scenes. Additionally, the explanation quality of answers was also increased via referring expression generation.

\section{Proposed Method}
\begin{sidewaysfigure*}
\centering
\includegraphics[width=1\linewidth]{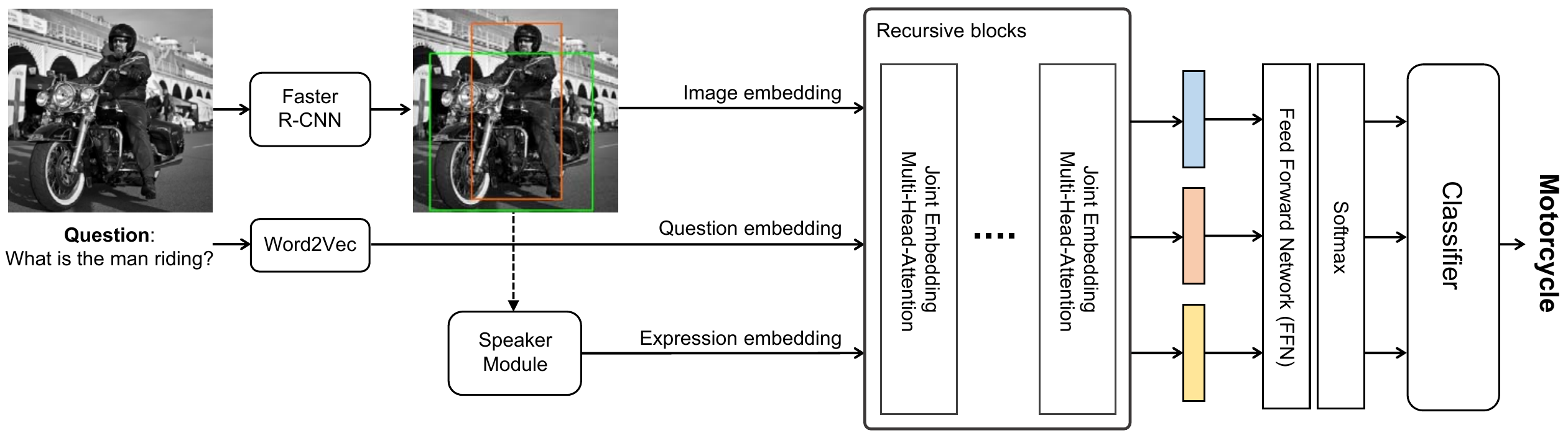}
\caption{Overall framework of the proposed VQA system. The JE-MHA network takes RoI pooled image embeddings, question embeddings, and generated expressions as input features. The model simultaneously learns multi-modal attention through a recursive structure and a classifier predicts final answers.}
\label{fig:framework}
\end{sidewaysfigure*}

We present an explainable VQA framework with local-scene-aware referring expression generation. The overall framework is illustrated in Fig. \ref{fig:framework}. From a given image and question, the speaker module generates referring expressions for each object considering the context of the target object. Three different embedding modalities, namely region of interest (RoI) pooled image embeddings, question embeddings, and expression embeddings are used as inputs for the VQA model. A combination of MHA modules comprises the JE-MHA network, which operates recursively to obtain three final feature vectors. Finally, the answer predictor computes a confidence-level for each answer candidate.

\subsection{Speaker module}
In this section, we present our approach to generating object expressions that are more detailed and precise for images with crowded backgrounds. The proposed speaker module considers neighboring features to understand the target object’s context. The referring expressions are generated for every object proposal of the object detector to bring as much additional knowledge as possible to answer the given question accurately and to provide a human-interpretable explanation. We also encode context features by calculating visual differences between a target and its neighboring objects.
We discuss local-scene-aware expression generation for each object. The speaker module can generate detailed and precise referring expressions for complex scenes.

\subsubsection{Local scene context with neighbors}
Pairs of objects in the same category, which are called positive pairs, have been frequently considered to generate unambiguous referring expressions in recent studies \cite{mao2016generation, liu2017referring, yu2017joint, tanaka2019generating, yu2016modeling}. The same category means that the pair of objects belongs to the same `supercategory’. In this paper, we use RefCOCO supercategories \cite{kazemzadeh2014referitgame}, which are semantic concepts encompassing several objects under a common theme. For example, the supercategory `vehicle’ contains `bicycle’, `car’, `bus’, `truck’, etc.
However, in real-world images, there are often many surrounding objects that are not in the same category. The relative locations or sizes of objects are frequently considered to identify positive pairs in an image, but these are insufficient for understanding the overall context of a scene. The context of the surrounding objects can be helpful for generating high-quality referring expressions.

In this regard, we propose a local-scene-aware referring expression generation method that considers neighboring objects that are associated with target object features.
Instead of utilizing only the locations and sizes of positive pairs of the target, differences in the visual features, locations, and sizes between the target and neighboring objects are extracted as visual comparison features. The explicit encoding of the visual difference between objects in the same category within an image is effective for the target context representation \cite{yu2016modeling}. Thus, we adopt the visual comparison technique to minimize ambiguity by learning the difference between the target and its positive pairs. 
This allows the proposed method to understand the discriminative aspects of the target object and leads to more precise expression generation, particularly in the scene with complex background characteristics.
For the object detector, we utilize the Faster-RCNN \cite{ren2015faster}, to detect objects and extract local-scene-aware features, including targets and neighbors.
We extract five features: the target object $o_{i}$, global context $g_{i}$ from the entire image, target location/size $l_{i}$, target context (differences in visual features) $\delta o_{i}$, and target location/size context (differences in location/size) $\delta l_{i}$.
The global context $g_{i}$ is modeled as the averaged features of objects detected by Faster-RCNN within a scene.
The target object $o_{i}$ is extracted as an RoI-pooled feature using a region proposal network \cite{anderson2018bottom}. The location and size of a target object in an image, which are defined by the $x, y$ coordinates of the top-left and bottom-right corners of the object, as well as its absolute size, are represented as $l_{i}=\left [ \frac{x_{tl}}{W}, \frac{y_{tl}}{H}, \frac{x_{br}}{W}, \frac{y_{br}}{H}, \frac{w_{i}\cdot h_{i}}{W\cdot H} \right ]$.

\begin{figure*}[!t]
\centering
\includegraphics[width=1\linewidth]{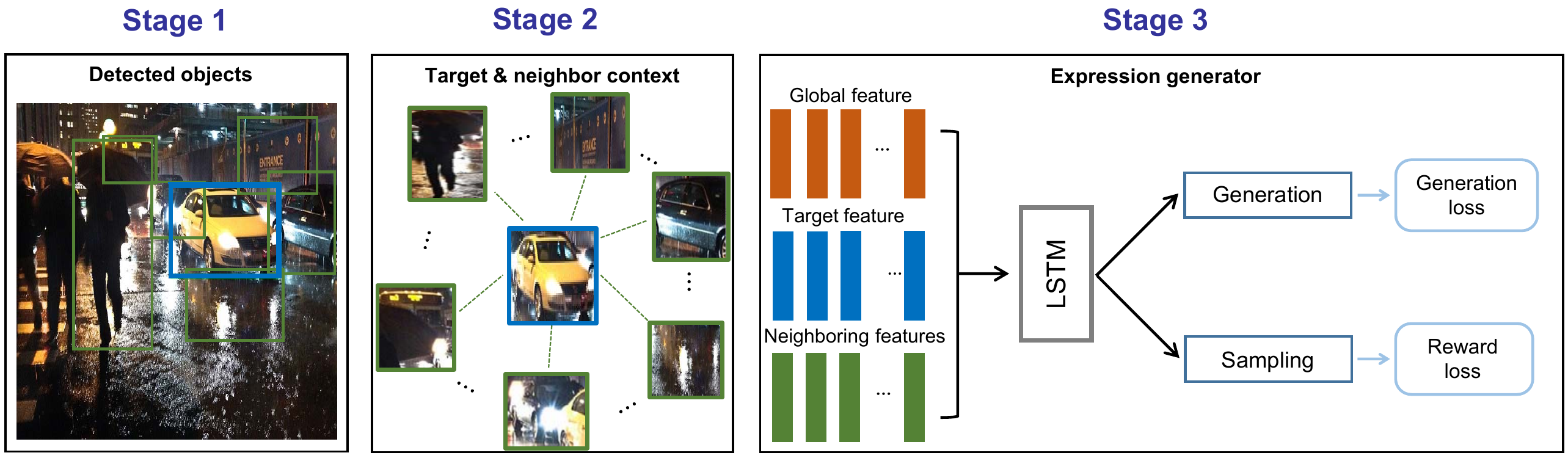}
\caption{High-level illustration of our speaker module. Each detected object constitutes a target object and its neighbors. The expression generator uses the global context, target object, and neighboring features to generate local-scene-aware text expressions. The LSTM is trained to minimize generation loss with reward loss.
}
\label{fig:my_label}
\end{figure*}

To identify meaningful neighbor objects around a target object, we first eliminate duplicate object detection results using non-maximum suppression (NMS).
After that, we select the $k$-nearest neighbor objects based on Euclidean distance to extract the local scene context.
Using the selected neighbors, we calculate the target context $\delta o_{i}=\frac{1}{k}\sum_{j\neq i}^{}\frac{o_{i}-o_{j}}{\left \| o_{i}-o_{j} \right \|}$, where k is the number of objects selected as neighbors and the target location/size context $\delta l_{i}=\left [ \frac{\left [ \bigtriangleup x_{tl} \right ]_{ij}}{w_{i}}, \frac{\left [ \bigtriangleup y_{tl} \right ]_{ij}}{h_{i}}, \frac{\left [ \bigtriangleup x_{br} \right ]_{ij}}{w_{i}}, \frac{\left [ \bigtriangleup y_{br} \right ]_{ij}}{h_{i}}, \frac{w_{j}\cdot h_{j}}{w_{i}\cdot h_{i}} \right ]$, which represents the relative location and size of the target object among the neighbor objects.

\subsubsection{Referring expression generation}
The local-scene-aware visual representation $v_{i}$ that is used to generate discriminative sentences is a combination of the five aforementioned image features and is defined as $v_{i}= W_{m}[o_{i}, g_{i}, l_{i}, \delta o_{i}, \delta l_{i}]$. To generate object expressions for each referred object, $x_t = [v_i; w_t]$ that is a concatenation of $v_i$ and a word embedding vector $w_t$ is fed into a long short-term memory (LSTM) to minimize the negative log-likelihood with the parameters $\theta$ as follows:

\begin{equation}
L_{s}^{1}(\theta) = -\sum_{i} log P(r_{i}|v_{i};\theta).
\label{eq:mle}
\end{equation}

We adopt the maximum mutual information \cite{mao2016generation} constraint that can reduce the ambiguity by minimizing the log-likelihood for positive pairs. Thus, the generated referring expressions for the target object, $o_i$, which is distinguished from the expressions for positive pairs.
There are two constraints; (1) the ground-truth expression $r_i$ should be more likely to be generated using the target object $o_i$ than using any other randomly sampled object $o_k$. (2) the target object should be more likely to generate the ground-truth expression $r_i$, rather than other expressions $r_j$ for positive pairs. The marginal loss $L_{s}^{2}(\theta)$ is calculated as follows:
\begin{align}
\nonumber L_{s}^{2}(\theta) = \sum_{i}{\lambda_{1}^{s}max(0, M_{1} + logP(r_{i}|v_{k})-logP(r_{i}|v_{i}))} \\ +{\lambda_{2}^{s}max(0, M_{2} + logP(r_{j}|v_{i}) - logP(r_{i}|v_{i}))},
\label{eq:margin_loss}
\end{align}
where $v_k$ denotes the visual representations of positive pairs of target objects, and $\lambda _{1}^{s}$, $\lambda _{2}^{s}$, and $M_{1}$, $M_{2}$ are marginal hyper parameters.
The optimized marginal loss indicates that the generated expression is more focused on the target objects than the positive pair. It is because the MMI constraint is designed to maximize and to minimize the log-likelihood for the target object and the positive pair, respectively.

We also adopt a reinforcer module \cite{yu2017joint} to generate precise referring expressions for each target object by calculating reward loss. Specifically, a two-layered multi layer perception (MLP) network is used to evaluate the consistency between generated expressions and visual features. Expression representations and visual representations are combined by the MLP to generate reward loss.
We use the local-scene-aware target object feature $v_i$ as a visual feature and employ an LSTM to encode the generated expressions as sentence features.
The evaluation scores are then used as rewards. The policy-gradient technique is used to optimize the reward function as follows:

\begin{equation}
{\bigtriangledown}_{\theta}J = -E_{P(W_{1:T}|v_{i})}[F(w_{1:T},v_{i}){\bigtriangledown}_{\theta}logP(w_{1:T}|v_{i};\theta)],
\label{eq:policy_gradient}
\end{equation}
where $w$ and $T$ denote a word vector and the number of words in a sentence, respectively.
The LSTM loss is used to minimize the expression of positive pairs while maximizing the expression of the target object by combining generation loss with reward loss.
Consequently, the speaker module can generate precise text expressions regarding the target object, which makes answers more accurate and increases the quality of the text expressions fed into the VQA.

\subsection{Image, question, and expression embedding}
In the proposed framework, the JE-MHA network takes three feature representations as inputs, namely; image embeddings, question embeddings, and expression embeddings. With regard to the image embedding, we use 36 detected object regions per image as a form of Up-down attention \cite{anderson2018bottom}.
A Faster R-CNN head \cite{ren2015faster} with a ResNet-101 network \cite{he2016deep} is employed for object detection.
The detection head was first pre-trained on the Visual Genome dataset \cite{krishna2017visual}, which was learned to detect 1,600 object categories and 400 attributes.
To generate an output set of image features $\textbf{X}$, we extract RoI pooled image features from the detected regions and perform NMS for each object category.
The Word2Vec method is applied for question embedding. 
We used a publicly available pre-trained model for the word embedding matrix of the 300-dimensional glove vectors and fine-tune weights during training \cite{pennington2014glove}.
Expression embedding is performed using a standard gated recurrent unit \cite{cho2014learning} with $512$ hidden units.
The hidden unit in the final stage is extracted as the embedding feature $x_c$ for the text expression of each object.

\subsection{Joint embedding multi-head-attention network}
\label{sec:MHA}
The visual, question, and expression embeddings are fed into the JE-MHA network module to generate fused features. In this section, we elucidate the working of the JE-MHA module, which consist of a combination of three MHA modules. Multiple JE-MHAs are utilized recursively, and final results are combined through a feed-forward network (FFN).

\begin{figure*}
\centering
\includegraphics[width=1\linewidth]{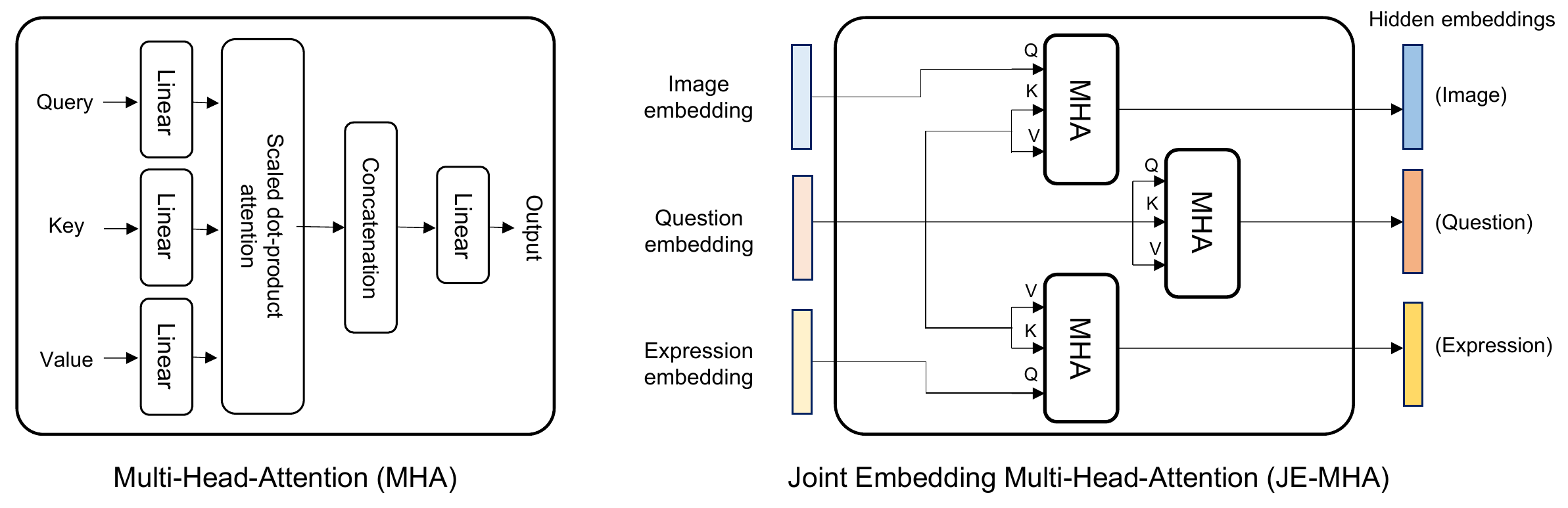}
\caption{(left) MHA module takes queries, keys and values as inputs. Outputs are computed as weighted sums of inputs. (right) JE-MHA network takes image, question, and expression embeddings as inputs. Each JE-MHA module outputs three different embeddings, which are used as inputs for the next module.}
\label{fig:mhanet}
\end{figure*}

\subsubsection{Multi-head-attention}
\label{sec:scaled_attention}
An attention function maps a query and set of key-value pairs to an output, as discussed in \cite{vaswani2017attention}. We compute the dot-product of the queries for all keys and divide each query by $\sqrt{d_k}$ to scale each query dimension. Then, the softmax function is applied to obtain weights for the individual values as follows:
\begin{align}
\text{Attention}(Q, K, V) = \text{softmax}(\frac{QK^{T}}{\sqrt{d_k}})\textbf{V}.
\label{eq:att}
\end{align}

The MHA is constructed by using linear projection to attend each query jointly based on key and value features in the same feature space. Then, a scaled dot-product is calculated $h$ times in parallel, as follows:

\begin{align}
\text{MHA}(\text{Q},\text{K},\text{V})=\text{Concat}(\text{head}_{1},...,\text{head}_{h})W^{O},\\
\text{head}_{i} = \text{Attention}(QW^{Q}_{i}, KW^{K}_{i}, VW^{V}_{i}).
\end{align}

The dot-product-based attention functions \cite{lu2019vilbert, yu2019deep, lu201912}, which consist of keys, queries, and values as input features are more effective at embedding two different input modalities compared to traditional attention techniques (e.g., additive attention) \cite{lu2016hierarchical, anderson2018bottom, teney2018tips}. Therefore, we utilize dot-product attention for our MHA to obtain more discriminative embeddings compared to additive attention techniques.

\subsubsection{JE-MHA network with image, question, and expression features}
\label{sec:vqa_mha}
The JE-MHA network utilizes three MHA modules to construct connections between image, question, and expression embedding features based on co-attention. Let $\textbf{Q} \in R^{N\times d}$ denote a set of question features containing $N$ tokens. Each token has $d$-dimensional features. Let $\textbf{X} \in R^{M \times d}$ denote the image embeddings of the $d$-dimensional features for $M$ detected objects. The attention module first utilizes the image embedding features $\textbf{X}$ to produce query vectors $q_x$.
It then uses the question features $\textbf{Q}$ to generate a key $k_q$ and value $v_q$.
The inner-product of the query $q_x$ and key $k_q$ are computed to learn the softmax result for the attention weights. The attention is then used to weight the value vectors $v_q$ to generate weighted features, as defined in Eq. (\ref{eq:att}).

In our multi-head setting, $h$ instances of the attended feature are computed using $h$ different projections of the original queries, keys, and values to improve the quality of attended features \cite{vaswani2017attention}.
The projected vectors have $d/h$ dimensions to ensure that the numbers of parameters are on the same scale.
The output of the JE-MHA module following six recursive processes based on the outputs of each MHA is a concatenation of the $h$-th attended features with dimensions of $M \times (d/h)$, resulting in a final output size of $F \in R^{M \times d}$.
Fig. \ref{fig:mhanet} presents the overall composition of the JE-MHA network and MHA module.

\subsubsection{Recursive block of the JE-MHA network}
\label{sec:recursive}
In the proposed VQA system, we combine JE-MHA modules recursively. The input embeddings $\textbf{Q}^i$, $\textbf{X}^i_o $, and $\textbf{X}^i_c$ are used as the question, image, and expression features for the $i$-th recursion, respectively. Additionally, we use $\text{MHA}$ and $\text{FFN}_d$ to denote an MHA module and the FFN layer with $d$ output units, respectively.
Based on the inputs from the $i$-th recursion, we compute the ($i+1$)-th recursion as follows:
\begin{align}
\textbf{Q}^{i+1} &= \text{MHA}( \textbf{Q}^{i}, \textbf{Q}^{i}, \textbf{Q}^{i}),\\
\textbf{H}_o^{i+1} &= \text{LayerNorm}(\text{MHA}( \textbf{X}^{i}_o, \textbf{Q}^{i}, \textbf{Q}^{i}) + \textbf{X}^{i}_o),\\
\textbf{X}^{i+1}_o &= \text{LayerNorm}(\text{FFN}_d(\textbf{H}_o^{i+1}) + \textbf{H}_o^{i+1}),\\
\textbf{H}_c^{i+1} &= \text{LayerNorm}(\text{MHA}( \textbf{X}^{i}_c, \textbf{Q}^{i}, \textbf{Q}^{i}) + \textbf{X}^{i}_c),\\
\textbf{X}^{i+1}_c &= \text{LayerNorm}(\text{FFN}_d(\textbf{H}_c^{i+1}) + \textbf{H}_c^{i+1}).
\end{align}

From the three final outputs of the recursive JE-MHA module, namely $\textbf{Q}^{t}$, $\textbf{X}^{t}_o$, and $\textbf{X}^{t}_c$, we construct a classifier to predict the confidence for each answer candidate. To compute the vector representations for $\textbf{Q}^{t}$, $\textbf{X}^{t}_o$, and $\textbf{X}^{t}_c$, we adopt the attention fusion approach as shown below:
\begin{align}
\textbf{q} = (\text{softmax}(FFN_1(\textbf{Q}^{t})))^T \textbf{Q}^{t}, \\
\textbf{x}_c = (\text{softmax}(FFN_1(\textbf{X}^{t}_c)))^T \textbf{X}^{t}_c,\\
\textbf{x}_o = (\text{softmax}(FFN_1(\textbf{X}^{t}_o)))^T \textbf{X}^{t}_o,
\end{align}
Finally, the answer confidence is computed as a projection of the joint question, image, and expression features as follows;
\begin{align}
A = FFN_{|dict|}(\textbf{q} \odot (\textbf{x}_c + \textbf{x}_o)),
\label{eq:ans}
\end{align}
where $|dict|$ denotes the word corpus of training data.

\section{Experiments}
We validated the effectiveness of the proposed method by comparing it to state-of-the art methods on the VQA v2.0 dataset \cite{antol2015vqa} to evaluate the accuracy of answer prediction and on the RefCOCO, RefCOCO+, and RefCOCO datasets \cite{kazemzadeh2014referitgame} to highlight the explainability of the proposed method.

\subsection{Datasets}
{\textbf{Visual question answering.}} We evaluated our proposed VQA model on the recent VQA v2.0 dataset \cite{antol2015vqa}. This dataset contains 1.1 million question and answer pairs based on MSCOCO images. The training, validation and test splits of the VQA 2.0 dataset were collected from the training, validation, and test splits of MSCOCO dataset, respectively.
This dataset includes complementary VQA pairs that ask the same question for two semantically similar images with different answers. 
Because these pairs of images are semantically similar, fine-grained reasoning is required to answer the question correctly. Therefore, our explanations focus on the important factors for predicting answers that can be compared for complementary pairs. Each question and expression have fixed lengths of 14 and 20 words, respectively.

{\textbf{Referring expression generation.}} We performed experiments on three expression datasets (RefCOCO, RefCOCO+, and RefCOCOg) \cite{kazemzadeh2014referitgame} collected from MSCOCO images. 
The RefCOCO and RefCOCO+ datasets are constructed based on the Image CLEF IAPR image retrieval dataset. For the RefCOCOg dataset, all the training, validation and test splits were constructed on images sampled from the training split of the MSCOCO dataset.
Since text expressions, which are generated from images, are utilized as an important component of the proposed method, we validated the quality of our generated text expressions. RefCOCO and RefCOCO+ were collected with time constraints for easily identifying target objects, whereas RefCOCOg was collected with no interactive settings and contains longer expressions than those in RefCOCO and RefCOCO+. Each dataset has 3.9 total objects and 1.63 objects of the same type per image. Overall, the three datasets contain 142,210, 141,565, and 104,560 expressions, respectively.

\subsection{Implementation details}
We optimized our speaker module using the Adam optimizer \cite{kingma2014adam} with a batch size of 128 and initialized the learning rate to 4e-4. The learning rate was set to decay by 0.5 every 500 iterations. The hidden state size of the LSTM generator for word embeddings was set to 512. Additionally, we empirically found that considering 20 neighbors with a 0.7 IoU in the NMS stage yields optimal results. For the reinforcer, our model generates three sample sentences and estimates the corresponding rewards. During the test, we used beam search with a beam size of 10.
For question and expression embedding, text preprocessing and tokenizing for UpDn features was performed using Stanford’s CoreNLP \cite{manning2014stanford} following \cite{anderson2018bottom}. The questions and expressions were first converted to lower case, trimmed to a maximum of $14$ words, then tokenized based on white spaces.
During the training phase, we trained our system on the VQA v2 training set for 13 epochs using the binary cross-entropy loss with soft scores, and BertAdam optimizer \cite{devlin2018bert} with a batch size of 64. After 10 epochs, the learning rate decayed by 0.2 every two epochs according to the method described in \cite{yu2019deep}.
For the evaluation of VQA tasks, during the training phase of the speaker module, we used the RefCOCOg dataset only to exploit their abundant expressions for the target object.

\subsection{Results for visual question answering with explanations}
In this experiment, we first evaluated the performance of our proposed VQA method on the VQA v2 dataset. We compared its performance with the state-of-the-art methods, i.e., Bottom-up \cite{anderson2018bottom}, MuRel \cite{cadene2019murel}, MCAN VQA \cite{yu2019deep}, Caption VQA \cite{wu2019generating}, MFH \cite{yu2018beyond}, MFH+Bottom-Up \cite{yu2018beyond}, BANGlCnt \cite{kim2018bilinear}, DFAF \cite{gao2019dynamic}, DCN \cite{nguyen2018improved}, and Counter \cite{zhang2018learning}.
After that, we also performed ablation studies to evaluate the contributions of input modalities and the effectiveness of using generated referring expressions.

\begin{table*}[]
\centering
{%
\begin{tabular}{|l|c|c|c|c|c|}
\hline
\multirow{2}{*}{Method} & \multicolumn{4}{c|}{Test-dev} & Test-std \\ \cline{2-6}
                        & All  & Yes/No  & Num & Other & All      \\ \hline
Bottom-up \cite{anderson2018bottom} & 65.32 & 81.82 & 44.21 & 56.05 & 65.67 \\
MuRel \cite{cadene2019murel} & 68.03 & 84.77 & 49.84 & 57.85 & 68.41 \\
Caption VQA \cite{wu2019generating} & 66.10 & 82.80 & 44.10 & 56.80 & 68.40\\
MCAN VQA \cite{yu2019deep} & 70.63 & 86.82 & 53.26 & 60.72 & 70.90 \\
MFH \cite{yu2018beyond} & 68.02 & - & - & - & - \\ 
MFH+Bottom-Up \cite{yu2018beyond} & 68.76 & 84.27 & 49.56 & 59.89 & - \\ 
BANGlCnt \cite{kim2018bilinear} & 70.04 & 85.42 & 54.04 & 60.52 & 70.35 \\ 
DFAF \cite{gao2019dynamic} & 70.22 & 86.09 & 55.32 & 60.49  & 70.34 \\ 
DCN \cite{nguyen2018improved} & 66.87 & 83.51 & 46.61 & 57.26 & 66.97 \\ 
Counter \cite{zhang2018learning} & 68.09 & 83.14 & 51.62 & 58.97 & 68.41 \\ 
JE-MHA (ours) & 69.70 & 86.30 & 50.80 & 59.90 & 70.79 \\ \hline
\end{tabular}%
}
\caption{VQA performance comparisons between the proposed method and competing methods on the VQA v2 test-dev and test-std split.}
\label{tab:VQAperf2}
\end{table*}

Table \ref{tab:VQAperf2} lists the answer prediction accuracies of each category and the corresponding average values on the VQA v2 test-dev split and test-standard split. 
We can see that the proposed JE-MHA VQA achieved 69.70\% and 70.79 \% accuracy on average, while the recent state-of-the-art method, MCAN VQA \cite{yu2019deep} achieves 70.63\% and 70.90\% on the test-dev split and test-std split of the VQA v2 dataset, respectively. This is because the proposed method showed a little short ability to count numbers. Even though the proposed method showed slightly lower accuracy on the test-dev split, it has the advantage of a capability of human-understandable explanation generation. We believe that the explainability can be the most promising characteristic of the proposed framework. Furthermore, the proposed method showed it showed one of highest accuracy in test-std split, and the state-of-the-art performance on the validation split, of the VQA v2 dataset.
We conducted a comparison with additional ablation studies on validation split of VQA v2 dataset to show the effect of text feature: We compared the JE-MHA VQA model in two different settings: 1) visual features only and 2) visual features + text features.

\begin{table*}[]
\begin{adjustbox}{width=1.0\textwidth}
\centering{%
\begin{tabular}{|l|c|c|c|c|c|c|}
\hline
Method & Visual & Textual & All & Yes/No & Num & Other \\ \hline
Tell-Answer \cite{qingli_tell_answer} & & \checkmark & 54.9 & 76.3 & 36.6 & 42.2 \\
JE-MHA VQA & & \checkmark & \bf{56.8} & \bf{76.5} & \bf{37.2} & \bf{46.9} \\ \hline
MCAN VQA \cite{yu2019deep} & \checkmark & & 66.4 & 84.3 & 48.2 & 57.7 \\
Caption VQA \cite{wu2019generating} & \checkmark & \checkmark & 65.8 & 82.6 & 43.9 & 56.4 \\
JE-MHA VQA (visdif+text) & \checkmark & \checkmark & \bf{66.7} & \bf{84.6} & \bf{48.3} & \bf{58.0} \\ \hline
\end{tabular}%
}
\end{adjustbox}
\caption{VQA performance comparisons with various input modalities on the VQA v2 validation split. ``Visual'' and ``Textual'' denote that the VQA models use visual or textual inputs, respectively.}
\label{tab:VQAperf}
\end{table*}

The experimental result demonstrates the effectiveness of each component of the proposed framework as shown in Table 2.
The textual feature without visual information showed 56.8\% on average, while the proposed model achieved 66.7\% with both visual and textual information. From this experiment, we can see that a baseline model using the generated expressions underperforms the proposed method. The results of the Tell-Answer method \cite{qingli_tell_answer} indicate that learning appropriate text expressions without visual features for VQA tasks can also yield good answer prediction performances. The proposed method also exhibits the best accuracy when only textual inputs are used. It is noteworthy that the proposed method not only provides an excellent answer prediction ability but also can provide explanations for the answers.

\begin{sidewaysfigure*}
\centering
\includegraphics[width=0.8\linewidth]{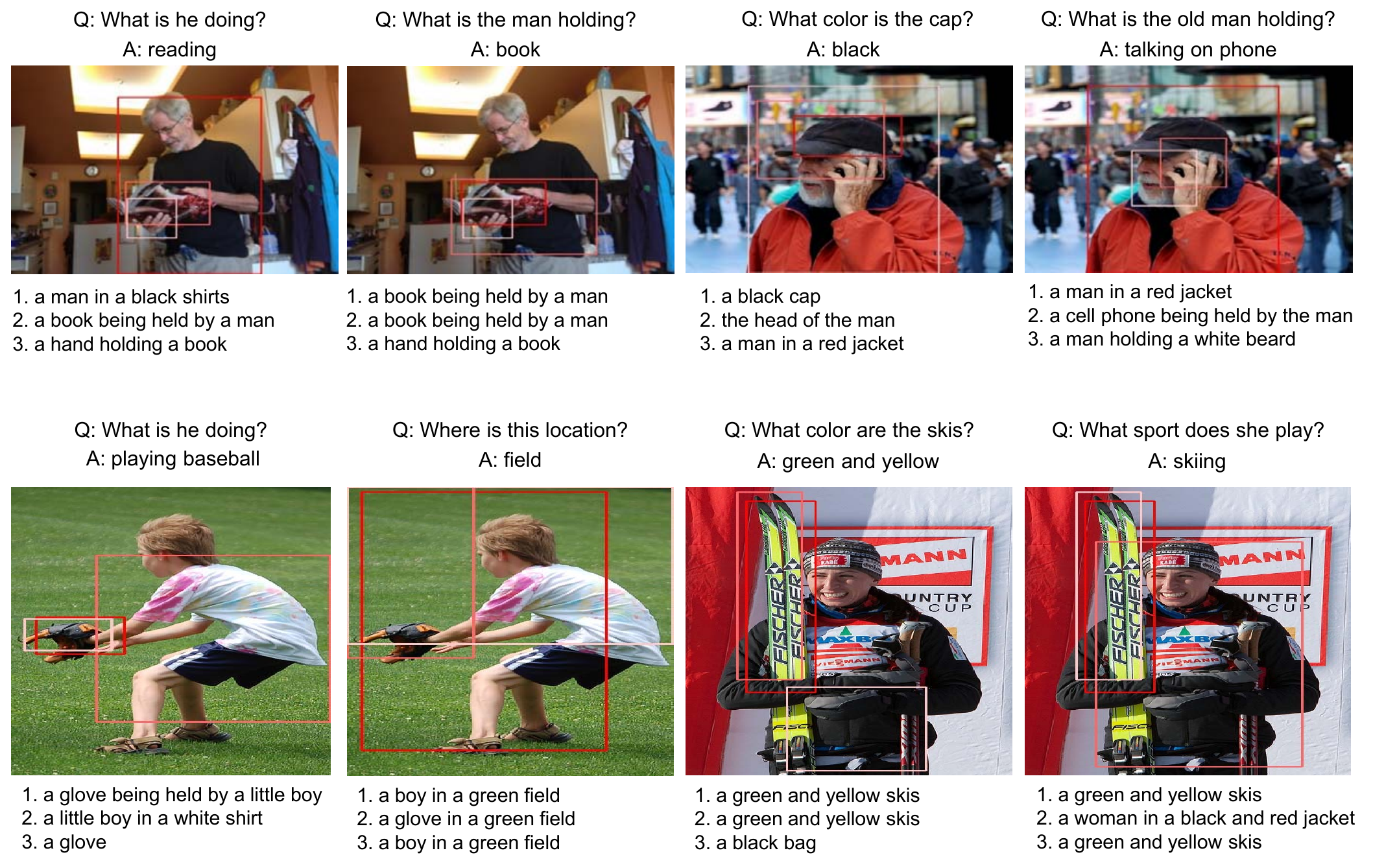}
\caption{Example VQA results with generated text explanations. The influential objects with the top three attention weights are indicated by bounding boxes. Each box is colored according to the importance of the corresponding visual feature, where darker red colors indicate higher weight.}
\label{fig:my_vqa_results}
\end{sidewaysfigure*}

\begin{figure*}[t]
\centering
\includegraphics[width=1\linewidth]{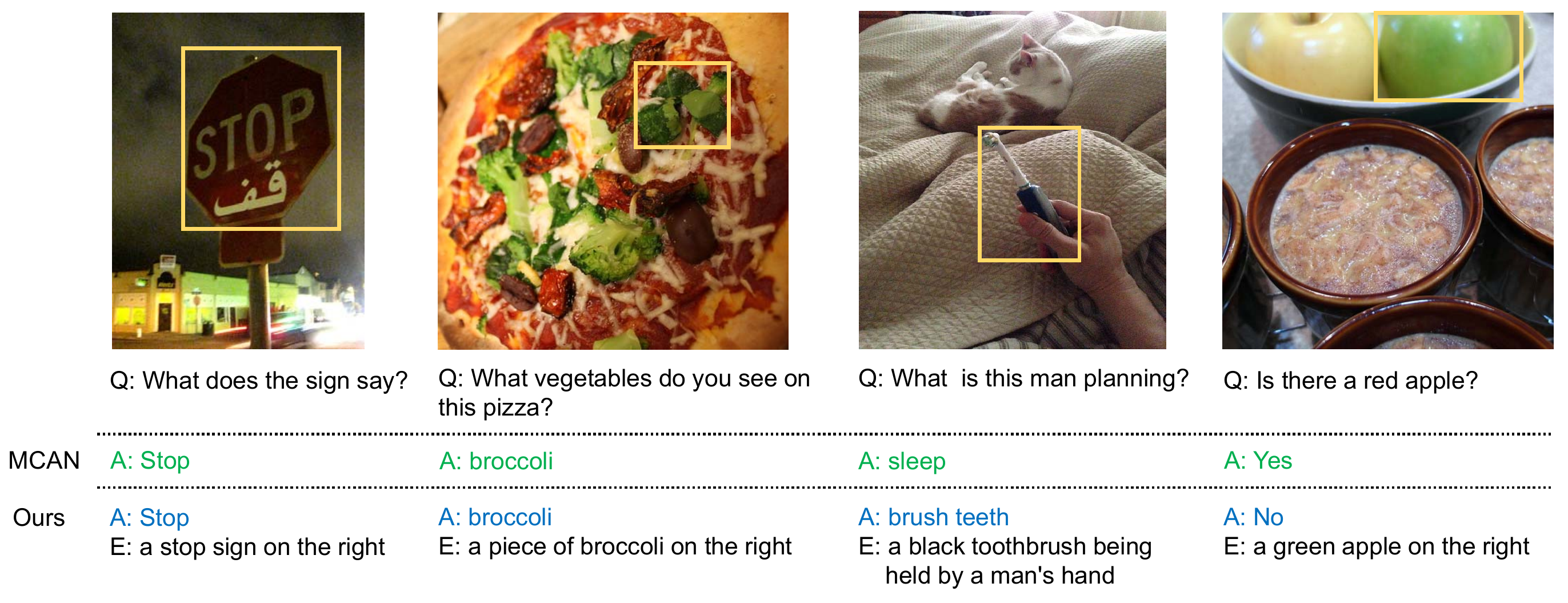}
\caption{Qualitative comparison to the MCAN method in terms of answer prediction. The yellow boxes indicate the most-attended objects in our model and the corresponding expressions are provided below (E:).}
\label{fig:qualitative_res}
\end{figure*}

Fig. \ref{fig:qualitative_res} presents comparisons between the proposed method and the MCAN \cite{yu2019deep} model. The green text in first lines represents the original MCAN predictions, the blue text in second lines represents the answers of our model, and the ``E’’ phrases in third lines represent the explanations generated from our model. Our approach not only yields superior performance, but also makes decisions understandable for humans by presenting natural language expressions.

Additional examples of answer predictions with generated explanations are presented in Fig. \ref{fig:my_vqa_results}. Two different questions are asked for each of the four images, and the information used for predicting answers is presented in order of importance.
The bounding boxes represent the most influential objects and their textual expressions with the top three attention weights. The generated human-readable sentences can also provide insights for interpreting the VQA model.
One can see that the bounding boxes of detected objects and generated explanations vary depending on the questions. The representational power of the speaker module is highlighted by the diverse expression generation in these examples.

\subsection{Results for referring expression generation}
Since the proposed method aims to generate human-understandable text explanations for the answer predictions, we evaluated the quality of generated explanations by comparing them to those provided by state-of-the-art referring expression generation methods, including the speaker-listener-reinforcer (SLR) model \cite{yu2017joint}, re-implemented SLR (re-SLR) \cite{tanaka2019generating} and referring expression based on grand theft auto $V$ (RefGTA) \cite{tanaka2019generating}.

\begin{sidewaystable*}[]
\textnormal
\centering
\resizebox{\textwidth}{!}{
\begin{tabular}{|l|c|c|c|c|c|c|c|c|c|c|c|}
\hline
\multicolumn{1}{|c|}{\multirow{3}{*}{Method}} & \multirow{3}{*}{\begin{tabular}[c]{@{}c@{}}Features \\ from.\end{tabular}} & \multicolumn{4}{c|}{RefCOCO} & \multicolumn{4}{c|}{RefCOCO+} & \multicolumn{2}{c|}{RefCOCOg} \\ \cline{3-12}
\multicolumn{1}{|c|}{} & & \multicolumn{2}{c|}{Test A} & \multicolumn{2}{c|}{Test B} & \multicolumn{2}{c|}{Test A} & \multicolumn{2}{c|}{Test B} & \multicolumn{2}{c|}{Val} \\ \cline{3-12}
\multicolumn{1}{|c|}{} & & Meteor & CIDEr & Meteor & CIDEr & Meteor & CIDEr & Meteor & CIDEr & Meteor & CIDEr \\ \hline
SLR \cite{yu2017joint}& VGGNet & 0.268 & 0.697 & 0.329 & 1.323 & 0.204 & 0.494 & 0.202 & 0.709 & 0.154 & 0.592 \\
SLR+rerank & VGGNet & 0.296 & 0.717 & 0.340 & 1.320 & 0.213 & 0.520 & 0.215 & 0.735 & 0.159 & 0.662 \\ \hline
re-SLR \cite{tanaka2019generating} & VGGNet & 0.279 & 0.729 & 0.334 & 1.315 & 0.201 & 0.491 & 0.211 & 0.757 & 0.146 & 0.679 \\
re-SLR+rerank & VGGNet & 0.278 & 0.717 & 0.332 & 1.262 & 0.198 & 0.476 & 0.206 & 0.721 & 0.150 & 0.676 \\ \hline
re-SLR \cite{tanaka2019generating} & ResNet & 0.296 & 0.804 & 0.341 & 1.358 & 0.220 & 0.579 & 0.221 & 0.798 & 0.153 & 0.742 \\ \hline
RefGTA SR \cite{tanaka2019generating} & ResNet & 0.307 & 0.865 & 0.343 & 1.381 & 0.242 & 0.671 & 0.220 & 0.812 & 0.164 & 0.738 \\
RefGTA SR+rerank & ResNet & 0.310 & 0.842 & \textbf{0.348} & 1.356 & 0.241 & 0.656 & 0.219 & 0.782 & 0.167 & 0.773 \\ \hline
RefGTA SLR & ResNet & 0.310 & 0.859 & 0.342 & 1.375 & 0.241 & 0.663 & 0.225 & 0.812 & 0.164 & 0.763 \\
RefGTA SLR+rerank & ResNet & 0.313 & 0.837 & 0.341 & 1.329 & 0.242 & 0.664 & 0.228 & 0.787 & 0.170 & 0.777 \\ \hline
Our SR & ResNet & 0.119 & 0.230 & 0.100 & 0.162 & 0.152 & 0.316 & 0.114 & 0.398 & 0.086 & 0.223 \\
Our SR+rerank & ResNet & 0.163 & 0.303 & 0.136 & 0.191 & 0.176 & 0.405 & 0.124 & 0.442 & 0.092 & 0.258 \\ \hline
Our SLR & ResNet & 0.322 & 0.905 & 0.342 & \textbf{1.393} & 0.260 & 0.722 & 0.235 & \textbf{0.853} & 0.177 & \textbf{0.896} \\
Our SLR+rerank & ResNet & \textbf{0.324} & \textbf{0.905} & 0.346 & 1.362 & \textbf{0.276} & \textbf{0.769} & \textbf{0.235} & 0.832 & \textbf{0.177} & 0.854 \\ \hline
\end{tabular}%
}
\caption{Comparison of our results to the state-of-the-art methods for referring expression generation for the RefCOCO, RefCOCO+, and RefCOCOg datasets. ``+rerank’’ indicates re-ranking the generated expressions according to the listener module. ``SLR’’ denotes the original SLR model and ``re-SLR’’ is a re-implemented version that uses ResNet image features as inputs.}
\label{tab:my-table}
\end{sidewaystable*}

The experimental results in Table \ref{tab:my-table} demonstrate that a reranking mechanism combined with a listener module improves expression quality in general, but such a combination is not very effective for the RefGTA model.
In particular, the speaker-reinforcer (SR) model yields worse performance than the SLR model. However, in our expression generation module, because our listener module can locate the referred objects accurately, better expressions are generated and our method outperforms all competing methods, including the RefGTA model.

\begin{figure*}[!t]
\centering
\includegraphics[width=1\linewidth]{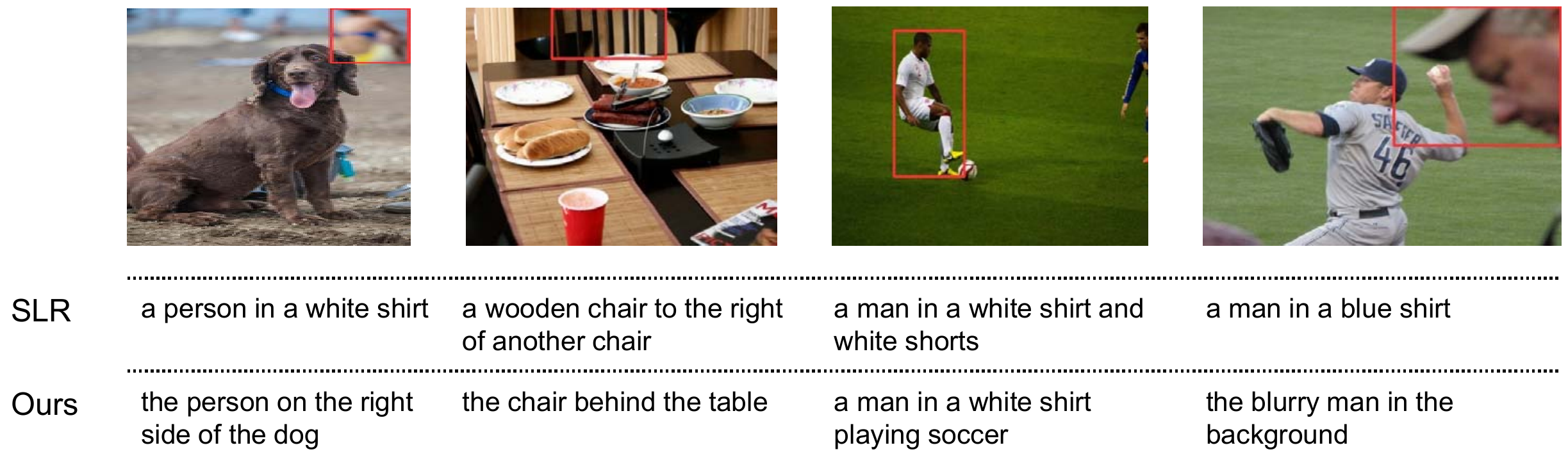}
\caption{Qualitative results for our referring expression generator on the RefCOCOg validation set and comparisons to the SLR model. Because the RefCOCOg dataset consist of longer sentences than RefCOCO, RefCOCO+, we trained our expression generator using the RefCOCOg dataset. }
\label{fig:refcocog}
\end{figure*}

For the qualitative comparison, we also compared our generated expressions to those generated by SLR \cite{yu2017joint} on RefCOCOg dataset, which was used to generate textual expressions for VQA.
Fig. \ref{fig:refcocog} presents the generated expressions and target objects.
One can observe that if the target object is occluded by other objects, our model understands the surrounding environment more accurately and generates less ambiguous sentences compared to SLR. This is important because our speaker module can generate referring expressions in complex situations, wherein ambiguous objects must be considered.

\subsubsection{Comprehensive evaluation}
In this section, we evaluate the performance of the speaker, listener, and reinforcer modules comprehensively to show the effects of each module for expression generation.
Our model is compared to the RefGTA \cite{tanaka2019generating} and SLR \cite{yu2017joint} models, which are based on SLR strategies, for evaluation. We calculated scores for the reinforcer/speaker modules using the ground-truth bounding boxes for all objects given $r$, $o^{*}=argmax_{i}F(r,o_{i})$, and $o^{*}=argmax_{i}P(r|o_{i})$.

\begin{table*}[p]
\textnormal
\centering
\begin{adjustbox}{width=1.0\textwidth}
\begin{tabular}{|l|c|c|c|c|c|c|c|c|}
\hline
\multicolumn{1}{|c|}{\multirow{2}{*}{Method}} & \multicolumn{2}{c|}{RefCOCO} & \multicolumn{2}{c|}{RefCOCO+} & RefCOCOg \\ \cline{2-6}
\multicolumn{1}{|c|}{} & Test A & Test B & Test A & Test B & Val \\ \hline
SLR (ensemble) \cite{yu2017joint} & 80.08\% & 81.73\% & 65.40\% & 60.73\% & 74.19\% \\
re-SLR(ensemble) \cite{tanaka2019generating} & 78.43\% & 81.33\% & 64.57\% & 60.48\% & 70.95\% \\ \hline
re-SLR (listener) & 81.14\% & 80.80\% & 68.16\% & 59.69\% & 72.36\% \\
RefGTA SLR (listener) \cite{tanaka2019generating} & 79.05\% & 80.31\% & 65.75\% & 62.18\% & 73.39\% \\
Our SLR (listener) & \textbf{82.67\%} & 78.83\% & \textbf{75.80\%} & \textbf{63.69\%} & 78.35\% \\
RefGTA SR (reinforcer) & 80.44\% & \textbf{81.04\%} & 67.81\% & 58.97\% & 74.94\% \\
Our SR (reinforcer) & 82.17\% & 79.04\% & 74.88\% & 62.81\% & \textbf{78.41\%} \\ \hline
re-SLR (speaker) & 80.70\% & 81.71\% & 68.91\% & 60.77\% & 72.55\% \\
RefGTA SR (speaker) & 82.45\% & \textbf{82.00\%} & 72.07\% & 61.06\% & 70.35\% \\
RefGTA SLR (speaker) & 83.05\% & 81.84\% & 72.37\% & 59.13\% & 74.79\% \\
Our SR (speaker) & 67.98\% & 64.83\% & 61.04\% & 48.40\% & 63.85\% \\ 
Our SLR (speaker) & \textbf{83.21\%} & 78.55\% & \textbf{76.40\%} & \textbf{63.75\%} & \textbf{78.31\%} \\\hline
\end{tabular}
\end{adjustbox}

\caption{Comprehensive evaluations on the RefCOCO, RefCOCO+, and RefCOCOg datasets. ``Ensemble’’ indicates using both the speaker and listener or reinforcer. Our three modules (speaker, listener, reinforcer) exhibit improved performance in most cases compared to the state-of-the-art methods.}
\label{tab:comp}
\end{table*}

The results of our expression comprehension experiments are listed in Table \ref{tab:comp}. The listener module plays a more important role in our model than in other models. This is because our system considers the neighboring features of each target object. As a result, our method can improve the performance of the listener module by selecting a local-scene-aware context feature to generate unambiguous expressions.

\subsection{Effectiveness of adjusting top-down attention}
The proposed method also has an advantage in its ability to adjust top-down attention weights based on local-scene-aware context analysis. In addition to providing explanations, our model can focus and understand detailed images and questions.

\begin{figure}[t]
\centering
\includegraphics[width=0.8\linewidth]{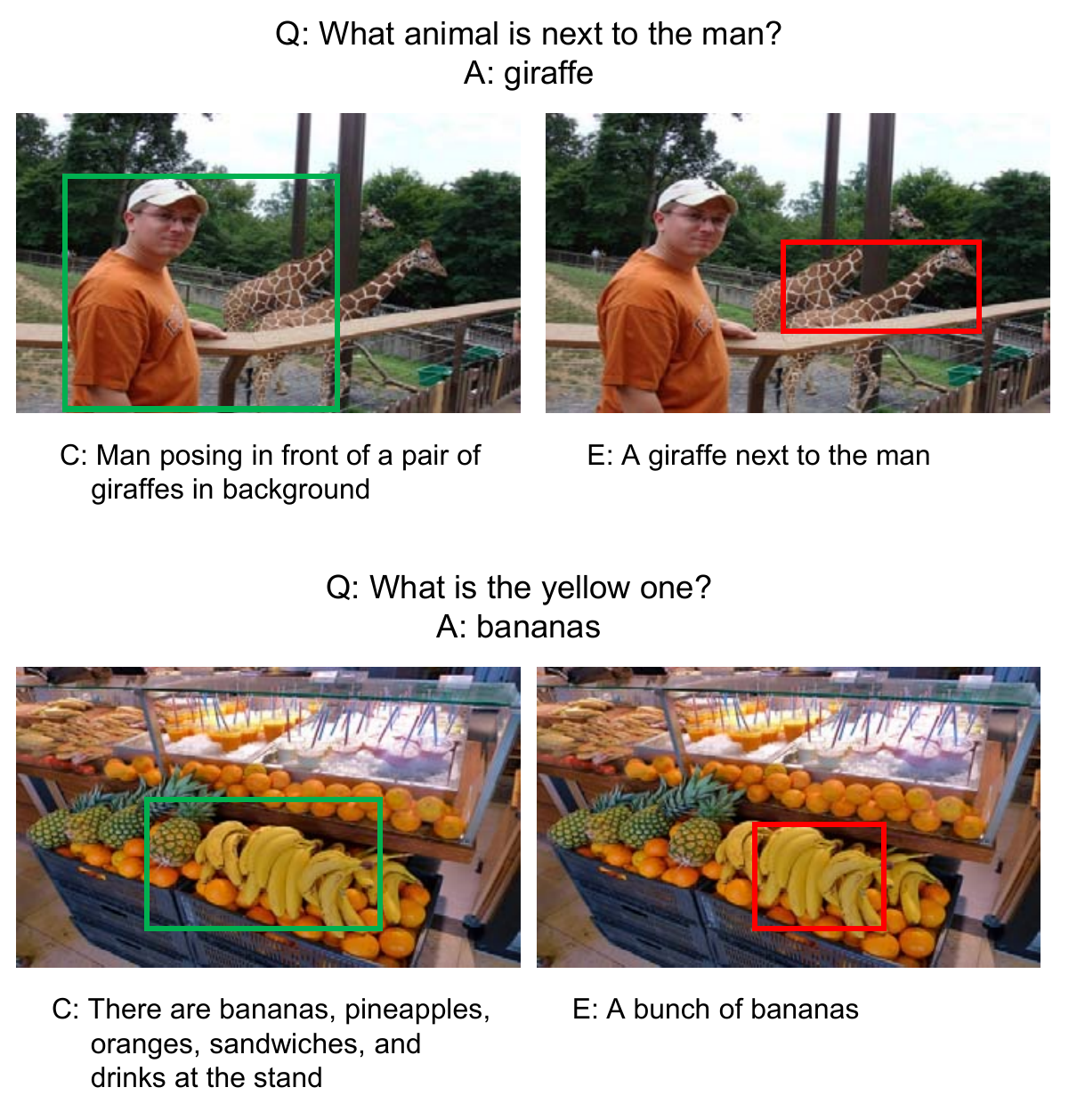}
\caption{Two examples of object-expression-based attention adjustment. The object expressions help the VQA model adjust visual attention more accurately. ``C’’ and ``E’’ denote image captions and explanations, respectively.
}
\label{fig:attn_adj}
\end{figure}

Fig. \ref{fig:attn_adj} demonstrates that our model can attend regions in complex scenes accurately. We compared our model to the Tell-Answer model \cite{qingli_tell_answer} to verify its effectiveness. Each box in the figure above represents the most attended region for predicting a final answer. In other words, the boxes represent the main focal areas of our model.
In the second example, our model also handles ambiguous references such as ``yellow one'' well.
One can see that using generated expressions helps our model accurately attend regions and understand questions in relatively complex VQA scenes.

\section{Conclusion}
We proposed a novel JE-MHA framework for visual question answering with explanations. The proposed speaker module generates local-scene-aware text expressions considering the spatial context of neighboring features. The resulting high-quality object expressions are combined with visual features and question embeddings through the proposed JE-MHA network.
Owing to representational power of text expressions, and co-attention of multimodal embedding, our VQA model can successfully predict answers for questions and provide rich explanations regarding scenes and questions.
Experimental results showed state-of-the-art performance on the VQA v2 dataset for answer prediction. We also showed expression quality using the RefCOCO, RefCOCO+, and RefCOCOg datasets by comparing different expression generation methods. Qualitative results for the generated expressions indicated that the proposed method can accurately answer questions when scenes are complex. In the future, we will extend our work on VQA with explanations to video data.

\section*{Acknowledgment}
This work was supported by Institute for Information \& communications Technology Planning \& Evaluation (IITP) grant funded by the Korea government (MSIT) (No. 2017-0-01779, A machine learning and statistical inference framework for explainable artificial intelligence, and No. 2019-0-00079, Artificial Intelligence Graduate School Program, Korea University).

\bibliography{egbib}


\end{document}